\documentclass[lettersize,journal]{IEEEtran}
\usepackage{amsmath,amsfonts}
\usepackage{algorithmic}
\usepackage{algorithm}
\usepackage{array}
\usepackage[caption=false,font=normalsize,labelfont=sf,textfont=sf]{subfig}
\usepackage{textcomp}
\usepackage{stfloats}
\usepackage{url}
\usepackage{verbatim}
\usepackage{graphicx}
\usepackage{cite}
\usepackage[cmyk]{xcolor}
\usepackage{lineno,hyperref}
\begin{document}



\title{Distribution-aware Interactive Attention Network and Large-scale Cloud Recognition Benchmark on FY-4A Satellite Image}

\author{Jiaqing Zhang,
	Jie Lei,~\IEEEmembership{Member,~IEEE},
	Weiying Xie,~\IEEEmembership{Senior Member,~IEEE},
        Kai Jiang,
        Mingxiang Cao,
	Yunsong Li,~\IEEEmembership{Member,~IEEE}

\thanks{This work was supported in part by the National Natural Science Foundation of China under Grant 62071360 and supported in part by the Innovation Fund of Xidian University under Grant YJSJ2302. (Corresponding~authors: Weiying Xie and Jie Lei)
Jiaqing Zhang, Jie Lei, Weiying Xie, Kai Jiang, Mingxiang Cao, and Yunsong Li are with the State Key
Laboratory of Integrated Services Networks, Xidian University, Xi'an 710071,
China (e-mail: jqzhang\underline{ }2@stu.xidian.edu.cn; jielei@mail.xidian.edu.cn; wyxie@xidian.edu.cn; mingxiangcao@stu.xidian.edu.cn; ysli@mail.xidian.edu.cn).}}

\markboth{Journal of \LaTeX\ Class Files,~Vol.~14, No.~8, December~2023}%
{Shell \MakeLowercase{\textit{et al.}}: A Sample Article Using IEEEtran.cls for IEEE Journals}


\maketitle

\begin{abstract}
Accurate cloud recognition and warning are crucial for various applications, including in-flight support, weather forecasting, and climate research. However, recent deep learning algorithms have predominantly focused on detecting cloud regions in satellite imagery, with insufficient attention to the specificity required for accurate cloud recognition. This limitation inspired us to develop the novel FY-4A-Himawari-8 (FYH) dataset, which includes nine distinct cloud categories and uses precise domain adaptation methods to align 70,419 image-label pairs in terms of projection, temporal resolution, and spatial resolution, thereby facilitating the training of supervised deep learning networks. Given the complexity and diversity of cloud formations, we have thoroughly analyzed the challenges inherent to cloud recognition tasks, examining the intricate characteristics and distribution of the data. To effectively address these challenges, we designed a Distribution-aware Interactive-Attention Network (DIAnet), which preserves pixel-level details through a high-resolution branch and a parallel multi-resolution cross-branch. We also integrated a distribution-aware loss (DAL) to mitigate the imbalance across cloud categories. An Interactive Attention Module (IAM) further enhances the robustness of feature extraction combined with spatial and channel information. Empirical evaluations on the FYH dataset demonstrate that our method outperforms other cloud recognition networks, achieving superior performance in terms of mean Intersection over Union (mIoU). The code for implementing DIAnet is available at \url{https://github.com/icey-zhang/DIAnet}.
\end{abstract}


\section{Introduction}
\label{sec:intro}

\IEEEPARstart{C}{loud} recognition and early warning systems are integral to aviation support, contemporary meteorological forecasting, climate science, and economic infrastructure \cite{wang2019intercomparisons}. Meteorological satellite imagery, with its expansive observational scope and rapid detection capabilities, has facilitated access to vast, uninterrupted atmospheric data. This is instrumental for monitoring meteorological shifts and projecting climatic patterns. The process of cloud recognition as a pixel-level segmentation is divided into two primary tasks: the initial task is to detect the presence and ascertain the positions of clouds, then the subsequent task involves the discernment and classification of the cloud types to identify their specific categories. Pixel-level segmentation algorithms \cite{sun2022irdclnet, sun2023danet, zhang2023rpmg}, bolstered by the advancement of deep learning technologies, have made rapid progress in the field of natural image processing. They are capable of dividing images into fine-grained pixel-level regions, thereby identifying and classifying various objects and textures within the images. 
\begin{figure}[tpb]
	\centering
	\includegraphics[scale=0.85]{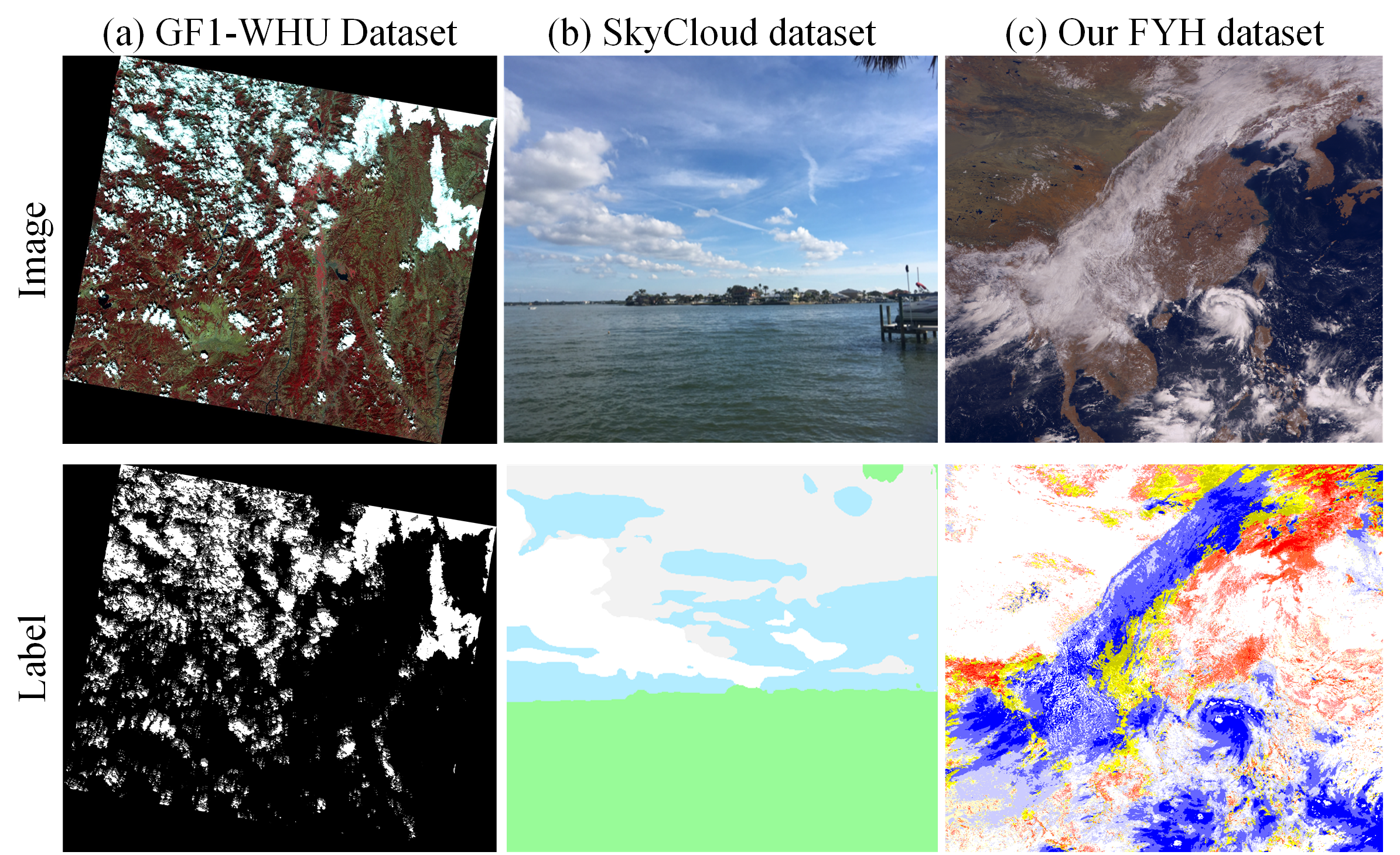}
	\centering
	\caption{Comparison of different datasets: GF-1-WHU dataset \cite{li2017multi} for satellite cloud detection, SkyCloud dataset \cite{gerhardt2023skycloud} for ground-based cloud detection, and our proposed FYH dataset for satellite cloud recognition.}
	\label{tasks}
 \vspace{-0.1in}
\end{figure}

However, cloud recognition in satellite imagery remains underexplored academically due to dataset complexity and the need for specialized algorithms. High-quality, annotated satellite datasets are scarce and expensive, demanding meteorological expertise for precise labeling. The task is further complicated by the high resolution, extensive coverage, and atmospheric influences of satellite images. Effective cloud identification algorithms must handle large, high-res data and distinguish clouds from different shapes.

In recent years, extensive research has been conducted on cloud detection across various satellite platforms, including Landsat \cite{LI2019197}, the Moderate Resolution Imaging Spectroradiometer (MODIS) \cite{platnick2003modis}, and Sentinel \cite{hagolle2010multi}. Cutting-edge deep learning algorithms \cite{zhang2021deep, zhang2021cloud, chen2023developing} have been thoroughly investigated for their efficacy in detecting clouds in satellite imagery. These algorithms demonstrate excellent capabilities in segmenting cloud boundaries and effectively detecting clouds, but they overlook the differentiation of cloud types \cite{li2022cloud} as shown in Fig. \ref{tasks} (a). The World Meteorological Organization (WMO) has established more fine-grained cloud classification criteria, categorizing clouds into ten genera and twenty-eight species \cite{world1987manual}. Distinguishing between these complex cloud formations presents a much more challenge than mere detection.

To date, many algorithms \cite{dev2015multi,taravat2014neural,zhuo2014cloud} have focused on processing ground-based cloud imagery, which simplifies the task of cloud shape recognition by categorizing specific cloud types or classifying based on thickness. The coverage of ground-based imagery is limited compared to the extensive reach of satellite imagery. However, the lack of precise category labels for clouds in satellite imagery presents a significant obstacle to developing systematic algorithms for cloud classification recognition in these images. 
Hence, constructing an effective cloud recognition benchmark and designing a viable and efficient recognition strategy is an urgent hot issue that needs to be addressed.

\begin{table*}[tpb]
	\small
	\renewcommand{\arraystretch}{1}
	\centering
	\setlength{\tabcolsep}{0.6mm}{
		\caption{Illustration of FYH cloud recognition dataset and existing satellite cloud detection datasets.}
		\label{dataset}
\begin{tabular}{cccccccccc}
\hline
\multicolumn{10}{c}{\textbf{Category A}: 0:clear 1:cloud 2:cloud shadow 3:shadow over water 4:snow/ice 5:water 6:land 7:flooded 8:thin cloud 9:thick cloud}  \\
\multicolumn{10}{c}{\textbf{Category B}: 0:Clear, 1:Ci, 2:Cs, 3:Dc, 4:Ac, 5:As, 6:Ns, 7:Cu, 8:Sc, 9:St, 10:Unknown} \\ \hline
\hline
Dataset   & Year & Satellite & Scene & Image number  & Category  & Channel & Resolution & Image Size  & Crop Size  \\
\hline
SPARCS \cite{hughes2014automated}     & 2014 & Landsat-8           & 80    &  - & A:1, 2, 3, 4, 5, 6, 7 & 10       & -          & $8000 \times 8000$ & $1000 \times 1000$ \\
GF1-WHU \cite{li2017multi}  & 2017 & GaoFen-1  & 108   & -  & A:0,1,2  & 4        & 16m        & -                  & -    \\
38-Cloud \cite{38-cloud-1} & 2019 & Landsat-8                              & 38 & 8,400Train/9,201Test   & A:1    & 4        & -          & $8000 \times 8000$ & $384 \times 384$   \\
95-Cloud \cite{38-cloud-1}   & 2020 & Landsat-8      & 95 & 3,4701Train/9,201Test  & A:1                                         & 4        & -          & $8000 \times 8000$ & $384 \times 384$   \\
KappaZeta \cite{domnich2021kappamask} & 2021 & Sentinel-2                             & 155   & -                    & A:0, 1, 2,
8       & 14       & -          & -                  & $512 \times 512$   \\
CloudSEN12 \cite{aybar2022cloudsen12} & 2022 & Sentinel-1,2                  & -     & 49,400                & A:0, 2, 8, 9                              & 13       & 10m        & -                  & $512 \times 512$   \\
CHLandsat8 \cite{du2023gated}         & 2023 & Landsat 8 & 64    & 22,616Train/10,080Test & A: 1                                                                  & 3        & -          & $8000 \times 8000$ & $352 \times 352$  \\ \hline
FYH  & 2023 & FY-4A,H08  & 70,419 & 110,000Train/5,500Test & B  & 14 & 4km & $1092 \times 2748$ & $100 \times 100$ \\
\hline
\end{tabular}}
\vspace{-0.1in}
\end{table*}

In this paper, we introduce a benchmark dataset specifically designed for cloud recognition tasks, named FYH as shown in Fig. \ref{tasks} (c). The FYH dataset incorporates FY-4A L1 Data as input and utilizes H08 Cloud-type Products for labeling. Employing a domain adaptive operation that accounts for projection, temporal resolution, and spatial resolution, we ensure precise alignment of the data. We delve into the intrinsic challenges of cloud recognition from satellite images by conducting a thorough analysis of cloud categories and the distribution of sample categories. Our novel high-resolution keeping branch is engineered to preserve high-resolution details throughout the network, which is particularly beneficial for retaining information about small-scale features that may span as little as a single pixel. Furthermore, our parallel multi-resolution cross branches are designed to extract more compact features across different resolutions and scales. This is achieved by cross-updating and fusing different branches to accurately depict the cloud outlines, capturing the inherent variability and diversity in cloud shapes. To contend with the uneven distribution across different categories, we employ a distribution-aware loss (DAL) that biases the network training towards more challenging samples. In addition, we incorporate an interactive attention module (IAM) to bolster the robustness of feature extraction, thereby enhancing the overall performance of cloud recognition.

The main contributions of this work are three-fold:
\begin{itemize}
 \item \textbf{Benchmark Production}: We believe this to be the first effort in constructing an FYH benchmark dataset with 115,500 aligned image data and labels. The dataset covers nine cloud categories specifically designed for cloud recognition tasks, filling a significant gap in the field.
\item \textbf{Problem Analysis}: Our work is the first to refine cloud recognition problems based on a detailed analysis of the characteristics and distributions of data, considering the complexity and variety of cloud formations.
\item \textbf{Algorithm Optimization}: We introduce a novel deep neural network, DIAnet, which effectively accumulates local and global context dependencies across different receptive fields. Our experimental results demonstrate that DIAnet outperforms other models in this domain. 
\end{itemize}

\begin{table}[tpb]
	\small
	\renewcommand{\arraystretch}{1}
	\centering
	\setlength{\tabcolsep}{0.6mm}{
		\caption{The information of the multispectral images between FY-4A L1 Data and H08 Cloud-type Product.}
		\label{datadifference}
		\begin{tabular}{ccc}
			\hline
			& FY-4A L1 Data & H08 Cloud-type Product   \\ \hline
			File Type      & HDF     & NetCDF     \\
			Projection     & NOM      & EQR     \\
			Observation Area    & China proper  &$60{}^\circ$S-$60{}^\circ$N,$80{}^\circ$E-$160{}^\circ$W    \\
			Temporal Resolution & 15min      & 10min      \\
			Spatial Resolution  & \begin{tabular}[c]{@{}c@{}}4km\\ (Pixel number:1092 \\ Line number:2748)\end{tabular} & \begin{tabular}[c]{@{}c@{}}5km\\ (Pixel number:2401,\\ Line number:2401)\end{tabular} \\ \hline  
	\end{tabular}} 
 \vspace{-0.1in}
\end{table}

\section{Benchmarks and Methodology}
\label{sec:format}
\subsection{Benchmarks}
In recent years, we have witnessed the rapid evaluation of cloud detection benchmarks including the SPARCS \cite{hughes2014automated}, GF-1-WHU dataset \cite{li2017multi}, 38-Cloud \cite{38-cloud-1}, 95-Cloud \cite{95-cloud}, KappaZeta \cite{domnich2021kappamask}, CloudSEN12 \cite{aybar2022cloudsen12} and CHLandsat8 \cite{du2023gated}. 
Table \ref{dataset} lists the profiles of these datasets, including satellite, scene, scale, categories, channels, resolution, and size. 
These datasets primarily focus on cloud detection and differentiate between thin and thick clouds based on cloud density. However, the lack of categories for different cloud morphologies hinders the broader application of these datasets to more complex, higher-level recognition tasks involving various cloud types.

Advanced Geostationary Radiation Imager (AGRI) is one of the main loads of the FY-4A stationary meteorological satellite. L1 Standard Data Record (SDR) Data of FY-4A/AGRI whose resolution is 4 km, describes the pre-processed product after quality control, earth navigation, and calibration from level 0 raw package data. Himawari–8 (H08) is a new geostationary meteorological satellite operated by the Japan Meteorological Agency (JMA). The Cloud types defined by the International Satellite Cloud Climatology Project (ISCCP) from cloud property are utilized as labeled masks for model training and testing. Each pixel in a cloud-type image is labeled as one of 11 classes with values ranging from 0 to 10 representing: 0-Clear, 1-Cirrus(Ci), 2-Cirrostratus(Cs), 3-Deep Convection (Dc), 4-Altocumulus (Ac), 5-Altostratus (As), 6-Nimbostratus (Ns), 7-Cumulus (Cu), 8-Stratocumulus (Sc), 9-Stratus (St) and 10-Unknown.

For cloud recognition for FY-4A satellite imagery, the cloud-type products from H08 are employed as the benchmark for performing validations. We produce a domain adaptive on projection, temporal resolution, and spatial resolution for the match consistency of the FY-4A L1 Data and H08 Cloud-type Product. To the best of our knowledge, this is the first time that such match consistency datasets have been used for the validation of a new development cloud recognition method.
\begin{figure}[tpb]
	\centering
	\includegraphics[scale=0.5]{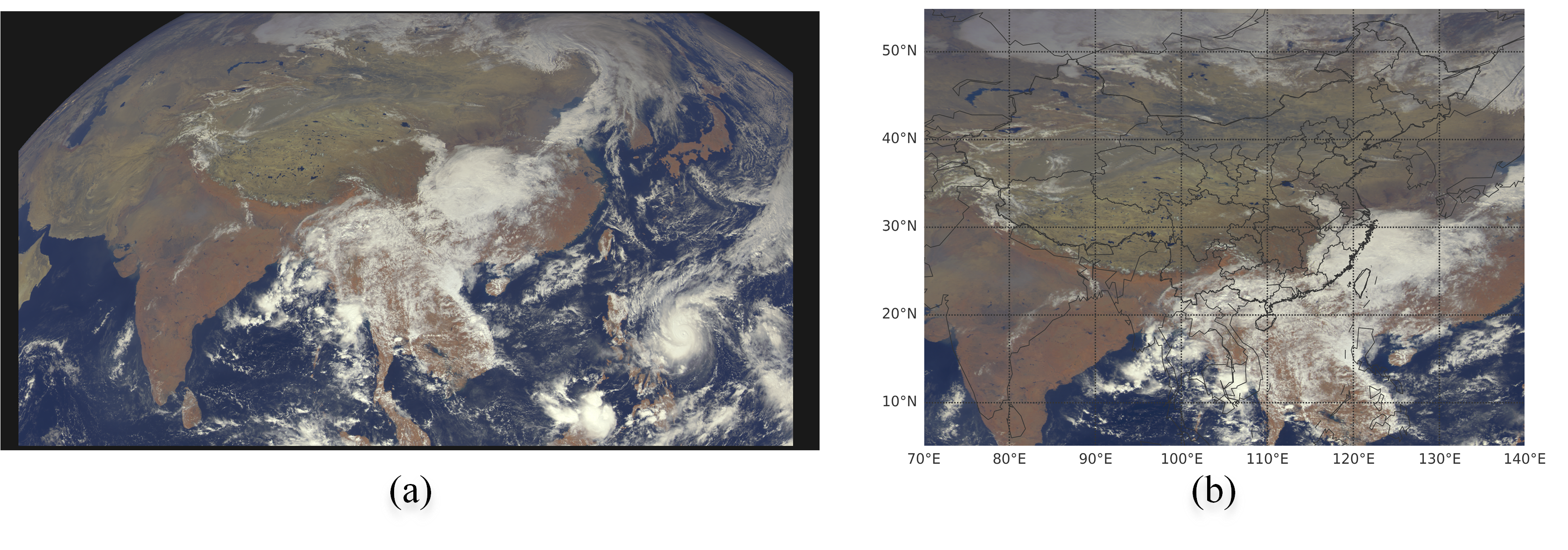}
	\centering
	\caption{Projection Transformation: (a) is NOM projection and (b) is the result of projection transformation to EQR projection.}
	\label{projection}
 \vspace{-0.1in}
\end{figure}

\begin{figure}[tpb]
	\centering
	\includegraphics[scale=0.35]{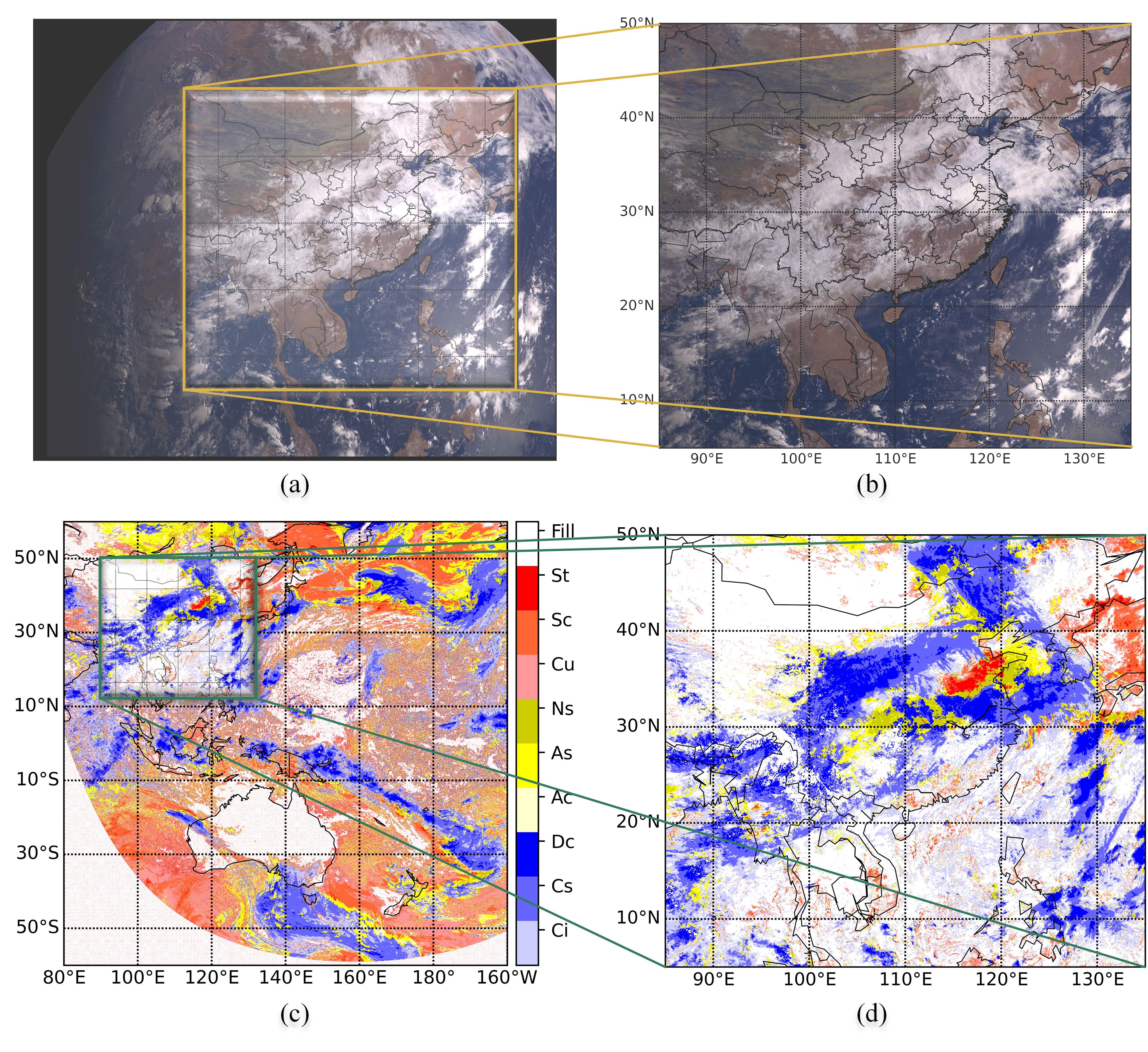}
	\centering
	\caption{Domain Adaptive: (a) FY-4A L1 Data, (c) H08 Cloud-type Product, (b) and (d) the result of domain adaptive. The class unknown and clear are merged into class fill for display.}
	\label{Match}
  \vspace{-0.1in}
\end{figure}


\subsection{Methodology}
\subsubsection{Domain Adaptation}
\label{DA}
We produce a domain adaptation on projection, temporal resolution, and spatial resolution for the aligned consistency of the FY-4A L1 Data and H08 cloud-type products. To eliminate the distinguish of the projection between FY-4A and H08, we transform the NOM projection of FY-4A to EQR (Equal-Rectangular) projection where the resolution is $0.05{}^\circ $ for space consistency for space consistency. Fig. \ref{projection} (b) and (d) is the result of projection adaptive from NOM projection ((a) and (c)) to EQR projection. We unify the common areas from $85{}^\circ $ E to $134.5{}^\circ $ E between $5{}^\circ $ N to $49.5{}^\circ $ N as input of model as illustrated in Fig. \ref{Match} (a) and (c). The data in the 2020 year is regarded as a research object. Furthermore, because of the unavailability of H08 cloud-type products at night in this area, the data only from 00:30 UTC to 10:50 UTC in a day is utilized and the time error between them is controlled within 2 minutes. For example, the data at 1:20 on July 18, 2020, of H08 cloud-type products showed in Fig. \ref{Match} (c) is matched as the label for FY-4A L1 data displayed in Fig. \ref{Match} (a) in which the observing beginning time is 1:19 on July 18. L1 Data of FY-4A is a pre-processed product after quality control, earth navigation, and calibration from level 0 raw package data. Hence, the range of values is normalized into $\left[ 0,1 \right]$ by maximum normalization without radiation calibration by bands. 
The Albedo of band01-band05 and the brightness temperature of band06-band14 is normalized individually.

\begin{figure}[tpb]
	\centering
\includegraphics[scale=0.5]{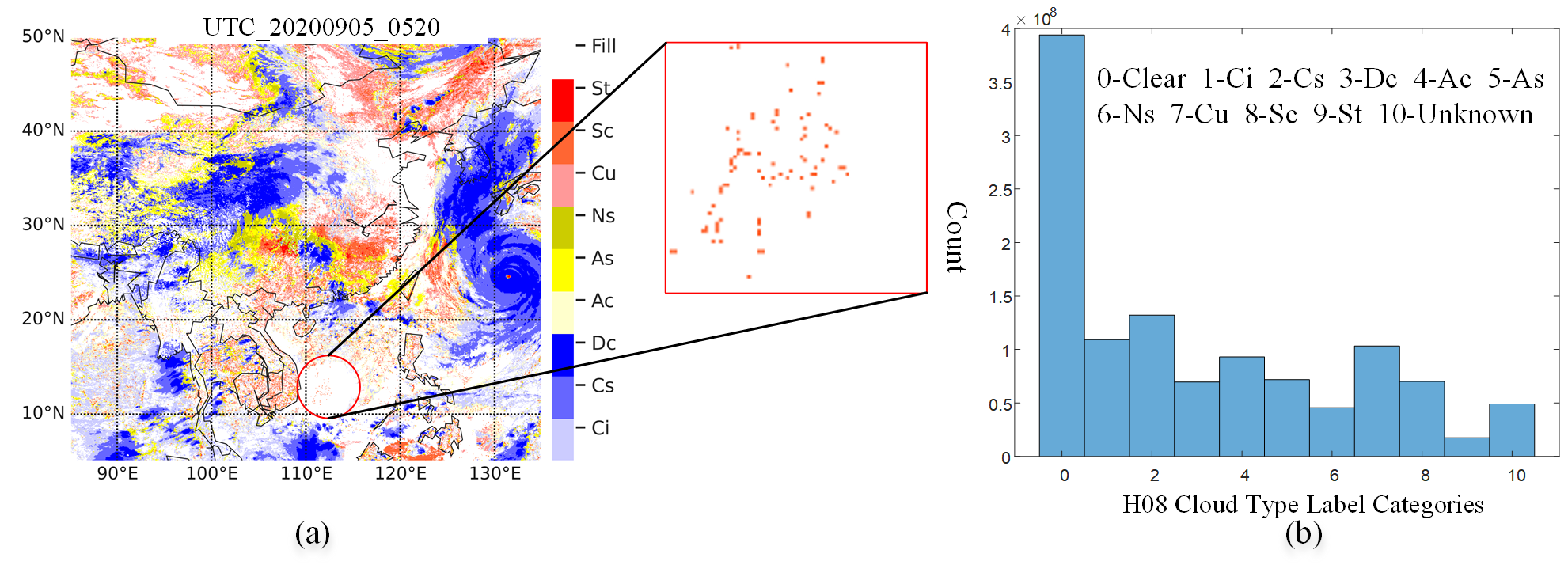}
	\centering
	\caption{An illustration of cloud categories and the distribution of sample categories, (a) H08 Cloud-type Product, (b) The cloud categories distribution}
	\label{data_distribution}
 \vspace{-0.1in}
\end{figure}

\subsubsection{Distribution of Sample Categories} 
Cloud Recognition aims to distinguish the cloud with various shapes and diverse types at the pixel level which is a hard problem limited by the cloud complexity. As shown in Fig. \ref{data_distribution}, data distribution analysis is completed to dig out the properties of data and provides a feasible and interpretable pertinence solution. Some issues can be found: (1) as shown in Fig. \ref{data_distribution} (a), some objects are small and even occupied in one pixel; once the resolution comes low, the object may be lost. (2) The cloud distribution is random and the cloud shape is not fixed; Owing to the variability and diversity of the cloud shape, it is difficult to delineate the cloud outlines. (3) shown in Fig. \ref{data_distribution} (b), the distribution of the categories is unbalanced, and the large amount of data in class 0-clear leads to the “long tail distribution” of the whole dataset and the learning process tends to undesired non-cloud data training.
\begin{figure}[tpb]
	\centering
	\includegraphics[scale=0.4]{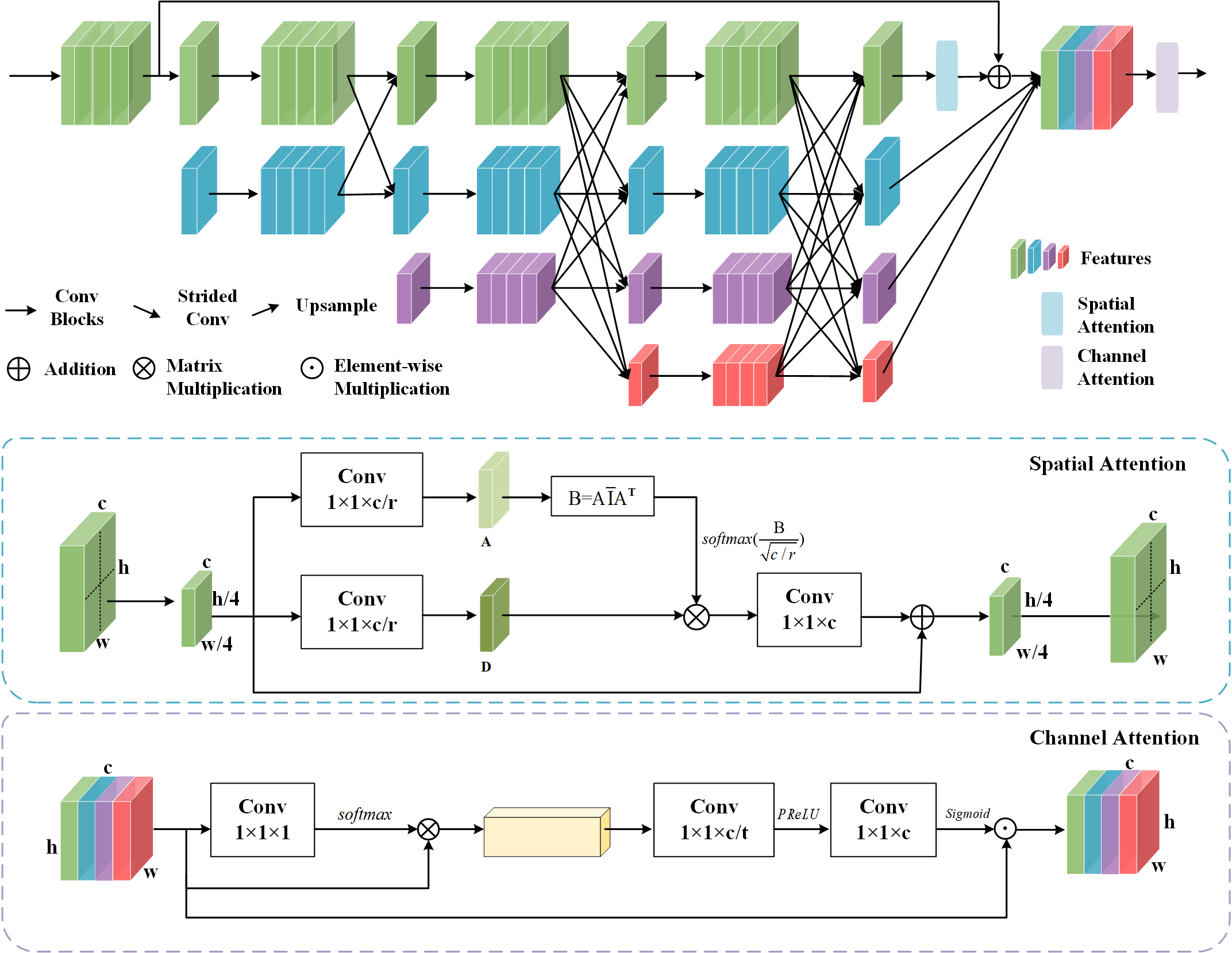}
	\centering
	\caption{An illustration of the proposed end-to-end DIAnet.}
 
	\label{Figure1}
 \vspace{-0.1in}
\end{figure} 
\subsubsection{DIAnet}
\label{DIAnet}
We have focused on designing an end-to-end data-driven network for cloud recognition to address the issues discussed above in the dataset. Consequently, we introduced a method called Distribution-aware Interactive-Attention Network (DIAnet). Fig. \ref{Figure1} illustrates a general overview of the DIAnet framework. Intuitively, the proposed DIAnet jointly trains branches in different resolutions in an end-to-end pattern. We adopted a similar approach to HRNet for instantiating the backbone \cite{HRnet}. The high-resolution keeping branch, as depicted in Fig. \ref{Figure1}, is designed to maintain a high-resolution representation throughout the network, which significantly aids in effectively addressing the issue (1) by preserving pixel-level details. The idea behind the four branches is to use features extracted at different scales, providing both local and global context. We arranged the four stages because early layers' feature maps retain more spatial details, resulting in refined boundaries, while deeper layers extract high-level semantic features. This allows us to obtain features from different receptive fields to overcome the issue (2). To mitigate the influence of issue (3), we introduced a simple yet remarkably effective OHEM algorithm \cite{OHEM} as our distribution-aware loss. This algorithm automatically selects hard examples and strengthens the training of these samples in unbalanced datasets. The interactive attention module, as shown in Fig. \ref{Figure1}, plays a crucial role in cloud recognition tasks. It can capture correlations between different spatial positions and channels and effectively fuse them. By introducing the interactive attention mechanism, we can better model the shape, texture, and context of clouds, thereby providing more comprehensive spatial-channel information.

\begin{table*}[htpb]
	\centering
	\caption{Comparison of recognition IoU results with different methods on the FYH dataset.}
	\label{result}
	\renewcommand{\arraystretch}{1.3}
	\setlength{\tabcolsep}{1.6mm}{
		\begin{tabular}{ccccccccccccc}
			\hline
			Method      & mIoU  & Clear & Ci    & Cs    & Dc    & Ac    & As    & Ns    & Cu    & SC    & St    & Unknown \\ \hline
			DeepLab v3+ \cite{chen2018encoder} & 35.21 & 60.92 & 38.81 & 56.28 & 56.45 & 17.78 & 31.78 & 35.29 & 21.66 & 32.24 & 33.09 & 3.02    \\
			Unet    \cite{ronneberger2015u}    & 36.64 & 58.91 & 40.35 & 58.66 & 60.42 & 19.07 & 33.54 & 39.94 & 20.10 & 30.86 & 35.04 & 6.09    \\
			Unet++    \cite{zhou2019unet++}  & 37.37 & 68.26 & 36.39 & 60.86 & 59.73 & 21.67 & 34.62 & 29.42 & 23.78 & 35.32 & 27.24 & 13.76   \\
			Unet+HFSA  \cite{he2020hybrid}  & 39.14 & 61.89 & 42.03 & 61.25 & 62.57 & 21.26 & 36.36 & 42.35 & 24.67 & 34.22 & 31.41 & 12.60   \\
			Gated-SCNN \cite{takikawa2019gated} & 41.03 & 62.61 & 44.30 & 62.96 & 64.29 & 24.61 & 38.86 & 44.30 & 26.45 & 34.24 & 34.34 & 14.36   \\
			DIANet      & \textbf{49.86} & \textbf{69.05} & \textbf{49.04} & \textbf{69.57} & \textbf{70.41} & \textbf{30.60} & \textbf{43.71} & \textbf{57.95} & \textbf{40.18} & \textbf{49.32} & \textbf{46.41} & \textbf{22.21}  \\ \hline
	\end{tabular}}
    \vspace{-0.1in}
\end{table*}

\begin{table}[htpb]
	\centering
	\caption{Effects of each component in our DIAnet.}
	\label{abl}
	\renewcommand{\arraystretch}{1.3}
	\setlength{\tabcolsep}{3mm}{
		\begin{tabular}{cc|cc}
			\hline
			Method    & mIoU    & Method               & mIoU                 \\ \hline
			HRNet     & 45.21 & HRNet+DAL & 47.73             \\
			HRNet+IAM         & 47.00 & DIANet    & 49.86             \\ \hline
	\end{tabular}}
 \vspace{-0.1in}
\end{table}

\begin{figure}[htb]
	\centering
	\includegraphics[scale=0.75]{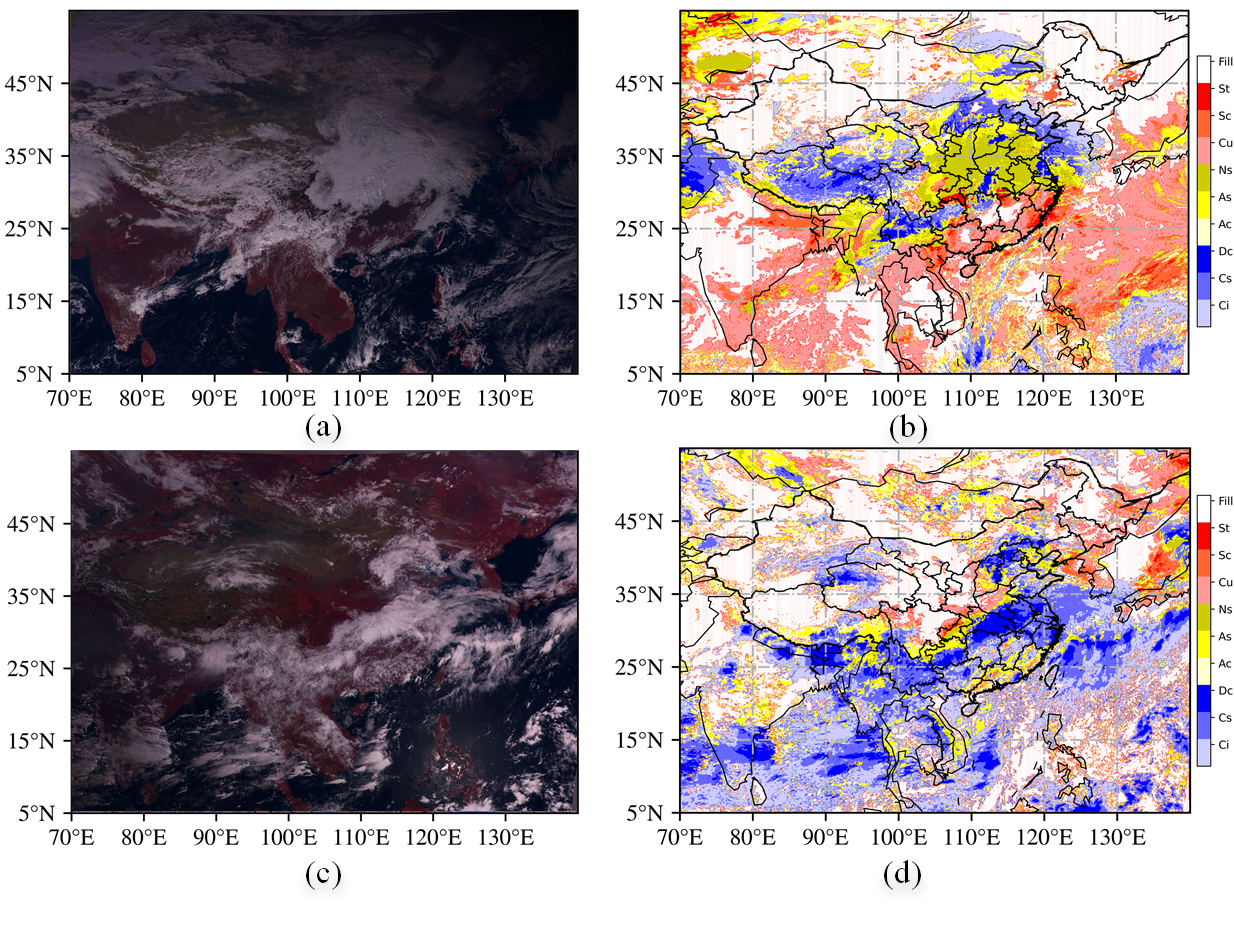}
	\centering
	\caption{The cloud-type products (b) and (d) generated by our proposed DIAnet on the FY-4A L1 EQR projection data (a) and (c).}
	\label{product}
 \vspace{-0.1in}
\end{figure} 

\section{Experience}
\subsection{Experience Setting}
\textbf{Dataset and Metrics.}  We generate our FYH dataset with FY-4A L1 Data and H08 cloud-type products as illustrated in Section \ref{DA}. After the screening, projecting, and cropping processing,  the 70419 data in the 2020 year are divided into a total of 110,000 training images and 5,500 validation images, with the size of $100 \times 100$. And we test the 20 complete images on the FY-4A/AGRI L1 EQR projection Data with size of $1000 \times 1400$ ($5{}^\circ $ N to $54.5{}^\circ $ N, $70{}^\circ $ E to $139.5{}^\circ$ E) to generate the final cloud-type products. We conduct the metric testing with size of $900 \times 1000$ (($5{}^\circ $ N to $49.5{}^\circ $ N, $85{}^\circ $ E to $134.5{}^\circ $ E)) aligned with the H08/AHI cloud-type label. The accuracy assessment measures the agreements and differences between the recognition result and the reference mask at the pixel level. The intersection over union (IoU) and mean IoU (mIoU) are used as metrics to evaluate the performance of the compared methods.


\textbf{Implementation Details.}  Our model is implemented using the Pytorch library. We utilize the SGD optimizer to train the model with a weight decay of 0.0005 and a momentum of 0.9. The number of epochs is set to 100 with the batch size of 16. We adopt the learning rate of 0.01 which is reduced by 10 times at epoch 30, 60, and 90. The whole training process is on 4 NVIDIA GeForce RTX 2080Ti GPUs. 
%


\subsection{Ablation Studies}
To understand how each module in DIAnet promotes cloud recognition accuracy, we test each component independently on the baseline HRnet and report the performance in Table \ref{abl}. The initial mIoU value starts from  45.21\% which is higher than other classical methods illustrated in Table \ref{result}. Such improvement indicates the necessity of the high resolution of the network features. The result gain of 2.52\% increase comes from DAL adaption which verifies the importance of the distribution-aware loss to overcome the issue that the distribution of the categories is unbalanced. When IAM is added to HRNet, the performance is promoted by 1.79\% mIoU. 
The final mIoU reaches 49.86\% when all three steps are adopted, bringing a 4.65\% absolute enhancement and validation of the effectiveness of our approach.

\subsection{Comparison with Other Methods}
As shown in Table \ref{result}, we compare DIAnet with other popular methods on our FYH dataset. The proposed method improves the performance by 9.23\% compared to the second-best method. We then visualize the cloud-type products generated by DIAnet. As shown in Fig. \ref{product}, our product can effectively sense cloud regions and obtain cloud category results at a pixel level with fine granularity. This demonstrates the importance of maintaining high-resolution features and extracting multi-scale features in high-resolution networks.

\section{Conclusion}
This paper explores the challenges in cloud recognition tasks and proposes solutions to address them. Specifically, we tackle the lack of cloud category datasets in satellite cloud imagery by constructing the FYH dataset and enabling deep learning networks to go beyond cloud detection and handle cloud recognition. Furthermore, our DIAnet model effectively handles issues such as small cloud targets and varying cloud shapes, imbalanced distributions between different categories through high-resolution branches, parallel multi-resolution cross branches, and incorporating a distribution-aware loss. Introduced interactive attention module enhances feature extraction robustness which results in a continuous improvement in cloud recognition accuracy. We validate the excellent performance of our method on the FYH dataset.
\ifCLASSOPTIONcaptionsoff
\fi
{
	\bibliographystyle{IEEEtran}
	\bibliography{refs}
}

\end{document}